\documentclass[10pt,twocolumn,letterpaper]{article}

\usepackage{wacv}
\usepackage{times}
\usepackage{epsfig}
\usepackage{graphicx}
\usepackage{amsmath}
\usepackage{amssymb}
\usepackage[accsupp]{axessibility}  
\usepackage{multirow}
\usepackage{pifont}

\def\wacvPaperID{****} 

\wacvfinalcopy 

\ifwacvfinal
\fi

\ifwacvfinal
\usepackage[breaklinks=true,bookmarks=false]{hyperref}
\else
\usepackage[pagebackref=true,breaklinks=true,colorlinks,bookmarks=false]{hyperref}
\fi

\begin{document}

\title{EdgeConv with Attention Module for Monocular Depth Estimation}

\author{
	Minhyeok Lee \quad
	Sangwon Hwang \quad
	Chaewon Park \quad
	Sangyoun Lee$^{*}$ \quad
	\vspace{0.01cm}\\
	Yonsei University \\
	{\tt\small \{hydragon516,sangwon1042,chaewon28,syleee\}@yonsei.ac.kr}
}

\maketitle

\ifwacvfinal
\thispagestyle{empty}
\fi

\begin{abstract}
	Monocular depth estimation is an especially important task in robotics and autonomous driving, where 3D structural information is essential. However, extreme lighting conditions and complex surface objects make it difficult to predict depth in a single image. Therefore, to generate accurate depth maps, it is important for the model to learn structural information about the scene. We propose a novel Patch-Wise EdgeConv Module (PEM) and EdgeConv Attention Module (EAM) to solve the difficulty of monocular depth estimation. The proposed modules extract structural information by learning the relationship between image patches close to each other in space using edge convolution. Our method is evaluated on two popular datasets, the NYU Depth V2 and the KITTI Eigen split, achieving state-of-the-art performance. We prove that the proposed model predicts depth robustly in challenging scenes through various comparative experiments.
\end{abstract}

\thispagestyle{empty}

\section{Introduction}

Monocular depth estimation is the task of generating a dense predictive depth map from a single image. The dense depth map provided with an RGB image is useful for understanding the three-dimensional geometric information of the scene in various computer vision tasks. In particular, it is critical to generate accurate depth maps for autonomous driving and robotics, where such information is essential. However, it is difficult to extract 3D structure information from a single monocular image. 

To solve this ambiguity, early depth estimation studies~\cite{saxena2005learning, malik2008novel, karsch2014depth} used hand-crafted methods based on probabilistic models. Recent approaches based on convolutional neural networks (CNNs) have achieved outstanding performance in numerous computer vision tasks. Owing to the success of deep learning in the computer vision field, many studies have adopted CNNs for monocular depth estimation. The latest studies~\cite{xu2018pad, qi2018geonet, zhang2018joint, lee2019big, chen2021attention} based on CNNs have demonstrated a better performance improvement than hand-crafted methods. However, these methods do not include a method to extract 3D structural information from a single 2D image. To extract depth information from a single monocular image, it is important to learn the structure information of the image. This structure information can be generated by learning the relationship between adjacent pixels, but few studies use this relationship or extract additional 3D information. Saxena \emph{et~al}.~\cite{saxena2005learning} introduced the first learning-based depth prediction using the relationship between different points in an image using a Markov random field (MRF). Gan \emph{et~al}.~\cite{gan2018monocular} proposed the affinity layer, representing the correlation of the values with the surrounding pixels, to extract relative features. Several studies~\cite{zhang2019pattern, xu2018pad, qi2018geonet} extracted structure information of robust images by generating surface normals as well as depth maps. However, using surface normals requires additional datasets, and the model performance is highly dependent on the surface normal accuracy.

\begin{figure}
	\setlength{\belowcaptionskip}{-24pt}
	\begin{center}
		\includegraphics[width=\linewidth]{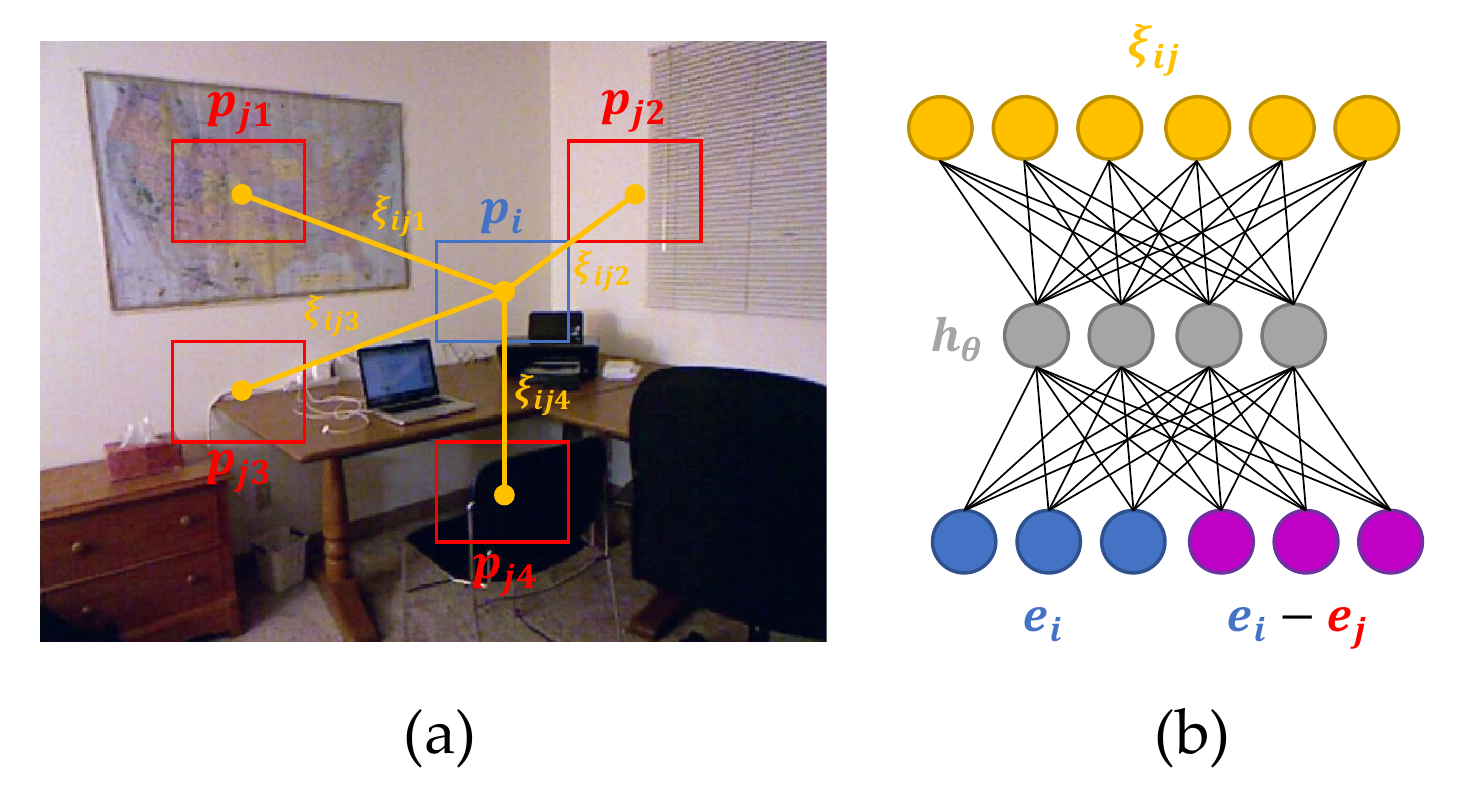}
		\caption{(a) Visualization of the proposed patch-wise edge convolution. (b) Details of patch-wise edge convolution (PEM). $e$ is the embedded patch $p$, $\xi$ is the edge features, and $h_{\theta}$ is a nonlinear function with learnable parameter $\theta$.}
		\vspace{-5ex}
		\label{fig:1}
	\end{center}
\end{figure}

In this paper, we propose a novel Patch-Wise EdgeConv Module (PEM) and EdgeConv Attention Module (EAM) to address the difficulty of generating structural information from a single image. Unlike the previous attention mechanisms, our proposed method extracts structural information based on the relationship between image patches using edge convolution. Edge convolution (EdgeConv) proposed by~\cite{wang2019dynamic} is based on graph convolution and learns structural information by extracting edge feature information between critical nodes of the 3D point cloud datasets. As shown in Figure~\ref{fig:1}, our PEM extracts edge features based on the relationship between adjacent patches in the feature space. To the best of our knowledge, our model is the first method to employ EdgeConv for depth estimation. The EAM generates a patch-wise attention map using edge feature information extracted by the PEM from multi-scale feature maps. To reconstruct 3D structure information from a single 2D image, it is essential to learn edge features representing the connection structure between adjacent pixels. The EAM combines the local structure information generated by the PEM and the global feature to make the model more robust to monocular depth estimation.

We conduct comprehensive evaluations on two challenging datasets, KITTI Eigen split~\cite{geiger2013vision} and NYU Depth V2~\cite{silberman2012indoor}. The results demonstrate that the proposed method achieves significant improvements over state-of-the-art methods, both qualitatively and quantitatively. Specifically, our method achieves the lowest root mean square error (RMSE) on both datasets.

Our main contributions can be summarized as follows:

\begin{itemize}
	\item We propose the novel PEM and EAM, which learns structural information between image patches by applying edge convolution.
	
	\item Our modules significantly improve the monocular depth estimation performance in challenging scenes by extracting edge features between patches close in feature space and learning the relationship between patches.
	
	\item The proposed network achieves state-of-the-art performance on the NYU Depth V2~\cite{silberman2012indoor} and KITTI~\cite{geiger2013vision} datasets. In addition, we prove the effectiveness of the proposed module through various comparative experiments.
\end{itemize}

\begin{figure*}[t]
	\setlength{\belowcaptionskip}{-10pt}
	\begin{center}
		\includegraphics[width=1\linewidth]{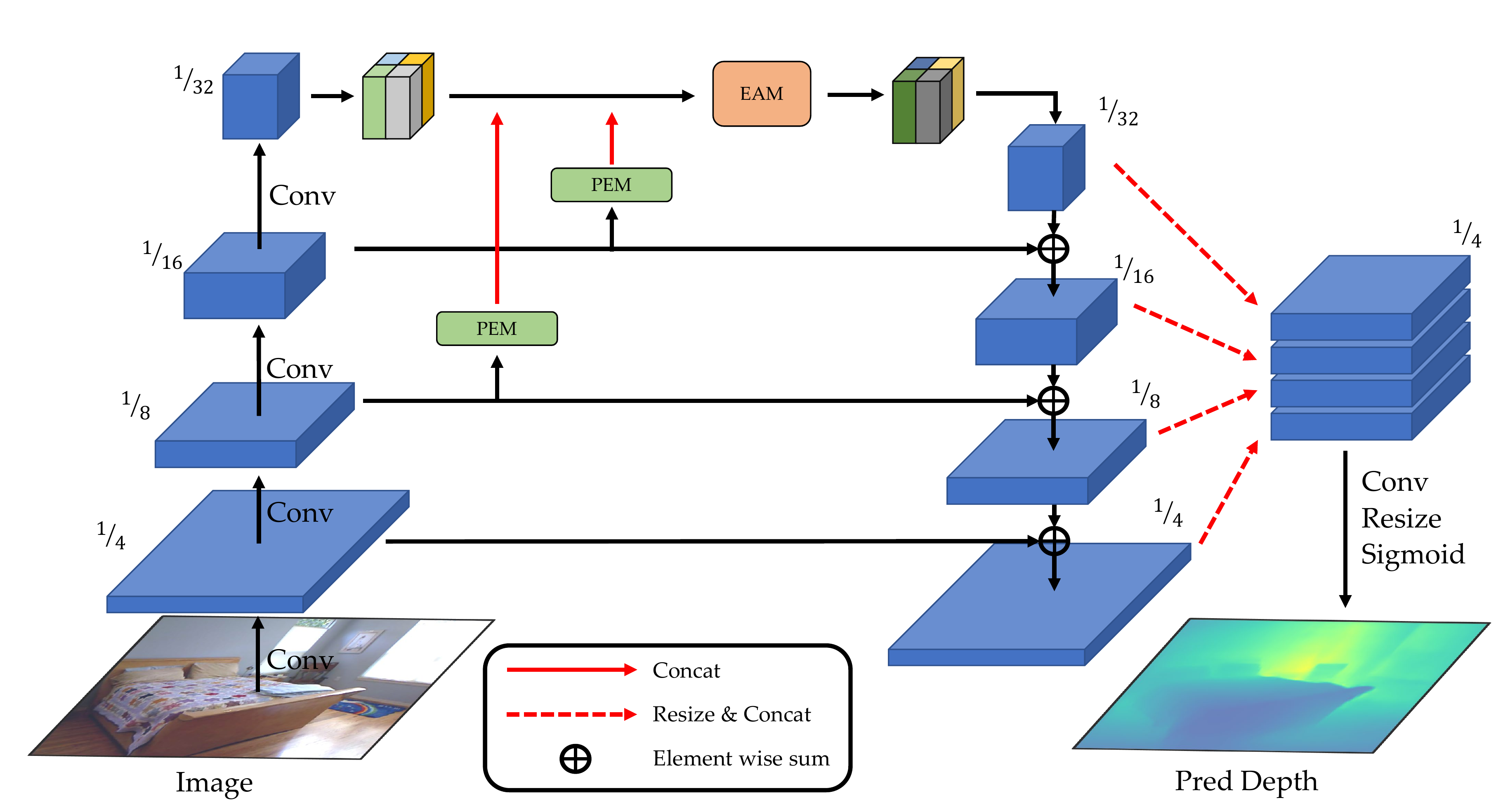}
	\end{center}
	\caption{Proposed architecture is an end-to-end framework based on the feature pyramid network (FPN)~\cite{lin2017feature}. The proposed PEM is located in the skip connection of the feature map layers of different sizes. The proposed EAM is located after the last layer of the encoder.}
	\label{fig:2}
\end{figure*}

\section{Related Work}

\noindent
\textbf{Supervised Monocular Depth Estimation.} Monocular depth estimation has been studied extensively. Saxena \emph{et~al}.~\cite{saxena2005learning} used discriminatively trained MRF to obtain a functional mapping of depth from visual cues. Liu \emph{et~al}.~\cite{liu2010single} used semantic segmentation and depth reconstruction to extract context information for depth estimation. Eigen \emph{et~al}.~\cite{eigen2015predicting} proposed a multi-scale convolutional architecture that predicted the coarse global depth and refined it progressively. Xu \emph{et~al}.~\cite{xu2018structured} applied Conditional Random Field to the model and used attention mechanisms to learn multi-scale features. Xu \emph{et~al}.~\cite{xu2018pad} proposed four multi-task learning methods for depth estimation: monocular depth prediction, surface normal estimation, semantic segmentation, and contour detection. Their paper argued that it could provide multi-modal information through multi-task learning. Qi \emph{et~al}.~\cite{qi2018geonet} proposed a mapping model between depth and surface normals to extract geometric information of images. Zhang \emph{et~al}.~\cite{zhang2018joint} introduced a task-recursive learning model that repeatedly performs depth estimation and semantic segmentation tasks. They also designed a task-attentional module to enhance the connectivity between the two tasks. Lee \emph{et~al}.~\cite{lee2019big} applied local planar guidance layers to the decoder to generate high-quality depth maps. Wang \emph{et~al}.~\cite{wang2020cliffnet} performed hierarchical embedding by applying depth-based scene classification and depth reconstruction. Chen \emph{et~al}.~\cite{chen2021attention} improved estimation performance by proposing a pixel-level attention model that effectively captured the pixel-level and image-level context.

\noindent
\textbf{Attention Mechanism.}
Attention mechanisms have provided significant improvements in machine translation and natural language processing. They have been used recently in various computer vision fields. Non-local neural networks~\cite{wang2018non} learn the relationships between pixels in different locations in the image. Fu \emph{et~al}.~\cite{fu2019dual} improved semantic segmentation performance using two types of non-local blocks. Zhang \emph{et~al}.~\cite{zhang2019self} introduced a better image generator with non-local operations. In addition, attention mechanisms can emphasize important spatial information of feature maps. For example,~\cite{park2018bam, woo2018cbam} meaningfully improved the classification task performance by adding channel attention and spatial attention mechanisms to the model.

\begin{figure*}[t]
	\setlength{\belowcaptionskip}{-10pt}
	\begin{center}
		\includegraphics[width=1\linewidth]{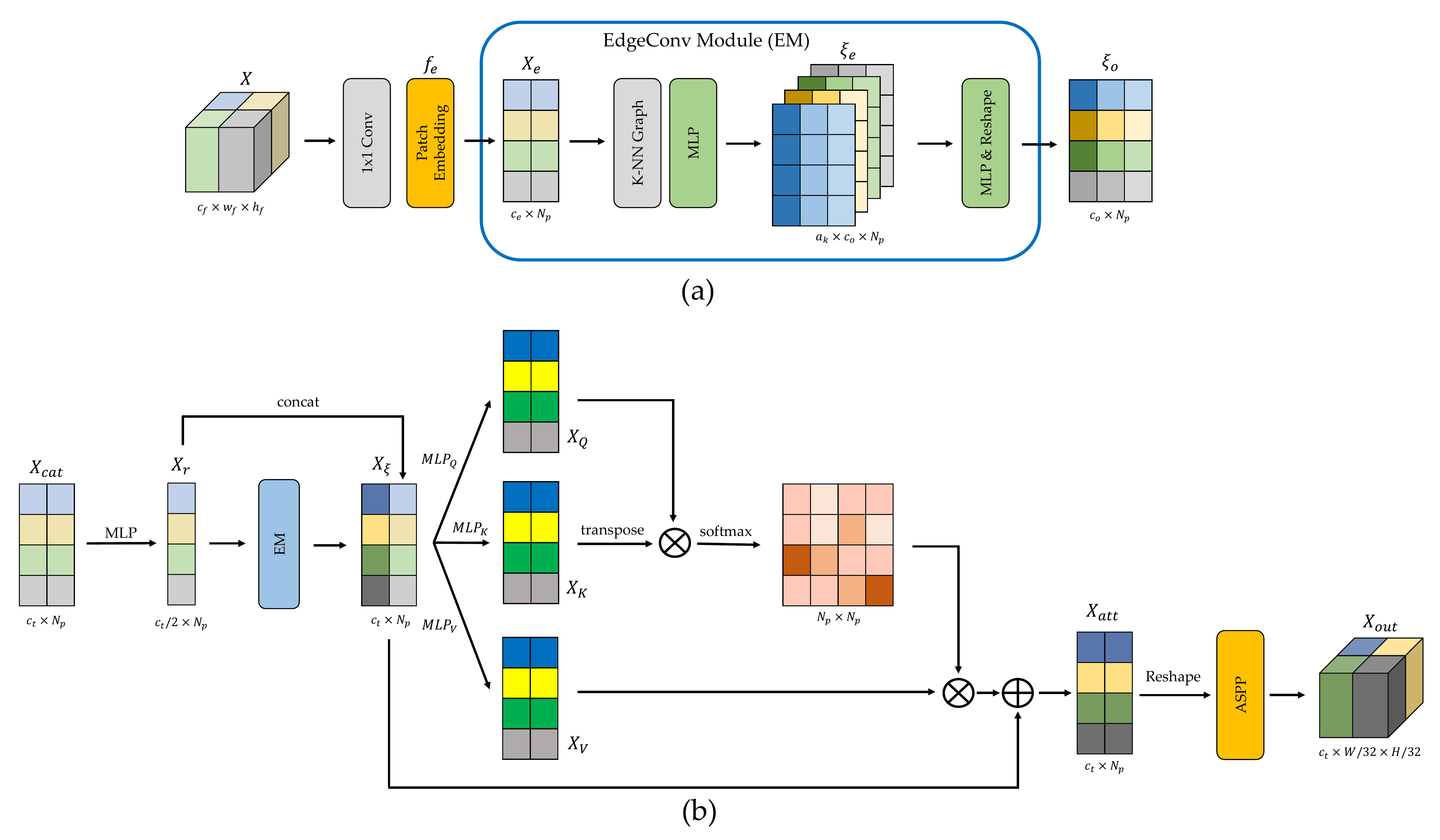}
	\end{center}
	\caption{(a) Structure of Patch-Wise EdgeConv Module (PEM). (b) Structure of EdgeConv Attention Module (EAM).}
	\label{fig:3}
\end{figure*}

\noindent
\textbf{Graph Convolutional Networks.}
Unlike CNNs, which learn features between image pixels, graph convolutional neural networks learn important features from connected nodes of the graph. Graph convolutional neural networks collect information from nodes and edges within the graph by applying convolution operations to the graph. Defferrard \emph{et~al}.~\cite{defferrard2016convolutional} introduced convolution for a graph as a multiplication for its spectral representation. Some studies~\cite{masci2015geodesic, simonovsky2017dynamic, bronstein2017geometric} focused on the convolution operation on the surface represented by the graph. In particular, graph convolution is often used to effectively obtain 3D structure information from point cloud data consisting of a large number of points. Representatively, DGCNN~\cite{wang2019dynamic} proposed EdgeConv, a type of graph convolution, to learn semantic displacement between key points and feature space neighbors.

\section{Proposed Approach}

\subsection{Overall Architecture}
The overall architecture is shown in Figure~\ref{fig:2}. Our model is based on the feature pyramid network (FPN) architecture~\cite{lin2017feature}. It includes two of the proposed PEMs and one EAM. The PEMs extract multi-scale edge features between patches of feature maps from the model encoder layers at different locations. The EAM performs patch-wise attention using the global depth contextual information of the last layer of the encoder and the multi-scale local edge feature. Finally, we resize and stack the decoder's multi-scale feature maps and pass them through the convolution and sigmoid layers. We multiply the output with the predefined maximum depth to get real depth values. In the following sections, we will cover the PEM and EAM in detail.

\subsection{Patch-Wise EdgeConv Module (PEM)}
\label{sect:3.2}
The PEM extracts patch-wise edge features corresponding to local structure information using EdgeConv~\cite{wang2019dynamic}. EdgeConv is designed for point cloud datasets, so it cannot be directly applied to 2D images. 

To solve this problem, the first step of the PEM is to divide the input feature maps into patches and generate linear embedding sequences of them. This process is the patch embedding proposed by~\cite{dosovitskiy2020image}. The patch embedding function $f_e$ is expressed as $f_e:p _ { i } \in \mathbb R^{c_f \times w_p \times h_p} \rightarrow e_i \in \mathbb R^{c_e}$, where $(w_p, h_p)$ is the patch size, $c_{f}$ is the number of channels of the feature map after the $1 \times 1$ convolution layer, $c_e$ is the length of the embedded vector, $p_{i}$ is the patch, and $e_{i}$ is the embedded patch at position $i$. Therefore, the size of the patch-embedded input feature map $X_{e}$ is $c_e \times N_p$, where $N_p$ is the number of patches. We set $N_p = \frac{H}{32} \times \frac{W}{32}$, where the height of the input image is $H$ and the width is $W$. The detailed reasons are covered in section~\ref{sect:3.3}.  

The second step of the PEM is to apply the EdgeConv Module (EM) as shown in Figure~\ref{fig:3} (a). The EM extracts edge features after generating graphs between patches using the k-Nearest-Neighbor (k-NN) algorithm. As shown in Figure~\ref{fig:1}, we define edge features as $\xi _ { ij } =h _ { \theta } \left ( e _ { i } , e _ { j } - e _ { i } \right )$, where $h _ { \theta } :\mathbb R ^ { c _ { e } } \times \mathbb R ^ { c _ { e } } \rightarrow \mathbb R ^ { c _ { o } }$ is a nonlinear function with a set of learnable parameters $\theta$ and $c_o$ is the length of the output edge feature $\xi _ { ij }$. This process is equivalent to generating dynamic graphs proposed by~\cite{wang2019dynamic}, and the output of this process is a tensor in the shape of $a_k \times c_o \times N_p$, where $a_k$ is the number of nearest neighbors. 

In the last step, the local edge feature map $\xi_o \in \mathbb R ^ {c_o \times N_p}$ is obtained by passing it through the channel-wise multi-layer perceptron (MLP) layer. As shown in Figure~\ref{fig:2}, we extract multi-scale edge features by applying the PEMs to two feature maps of different resolutions. The two edge feature maps are used as inputs to the EAM.

\subsection{EdgeConv Attention Module (EAM)}
\label{sect:3.3}
The EAM performs patch-wise attention from edge features generated from the PEM using a self-attention mechanism. The structure of the proposed EAM is shown in Figure~\ref{fig:3} (b). The EAM input is defined as follows:

\begin{equation} 
	X _ { cat } =\text{concat} \left ( f _ { g } ; \xi _ { o/8 } ; \xi _ { o/16 } \right )
\end{equation}

\noindent
When the height of the input image is $H$ and the width is $W$, $f_g \in \mathbb R^{c_g \times \left( \frac{W}{32} \times \frac{H}{32} \right)}$ is the spatially flattened feature map of the last layer of the encoder, where $c_g$ is the number of channels. $\xi _ { o/8 } \in \mathbb R^{c_{o/8} \times N_p}$ and $\xi _ { o/16 } \in \mathbb R^{c_{o/16} \times N_p}$ are the results of the PEMs for encoder feature maps with $1/8$ and $1/16$ size input images. Because we set $N_p = \frac{H}{32} \times \frac{W}{32}$, the size of $X_{cat}$ is $c_t \times N_p$, where $c_t = c_g + c_{o/8} + c_{o/16}$. 

The next step is to extract the edge features using the EM defined in section~\ref{sect:3.2}. This process is defined as follows:

\begin{equation} 
	X_{\xi} =\text{concat} \left ( X _ { r } ;\text{EM} \left ( X _ { r } \right ) \right )
\end{equation}

\noindent
$X_{cat}$ passes through the channel-wise MLP layer to become $X_r \in \mathbb R^{\frac{c_t}{2} \times N_p}$, which has half the number of channels of $X_{cat}$. Because we set the output channel of EM to $\frac{c_t}{2}$ as shown in Figure~\ref{fig:3} (b), the size of $X_{\xi}$ is the same as $X_{cat}$. 

The final process of the EAM is the MLP blocks and self-attention module (SAM). The self-attention module input is a triplet of $X_K$ (key), $X_Q$ (query), and $X_V$ (value), similar to~\cite{vaswani2017attention}. The three inputs of the SAM are expressed as $X _ { K } =\text{MLP} _ { K } \left ( X _ { \xi } \right ) ,X _ { Q } =\text{MLP} _ { Q } \left ( X _ { \xi } \right )$ and $X _ { V } =\text{MLP} _ { V } \left ( X _ { \xi } \right )$. The
self-attention formula is:

\begin{equation}
	\text{SAM} \left ( X _ { Q } , X _ { K } ,X _ { V } \right ) =\text{softmax} \left ( \frac{ X _ { Q } X _ { K } ^ { T } } { \sqrt { d } } \right ) \cdot X _ { V },
\end{equation}

\noindent
where $d$ is the channel of the SAM inputs. The final EAM output $X_{att} \in \mathbb R^{c_t \times N_p}$ is expressed as follows:

\begin{equation}
	X _ { att } =X _ { \xi } +\text{SAM} \left ( X _ { Q } ,X _ { K } ,X _ { V } \right )
\end{equation}

\noindent
In addition, $X _ { att }$ is reshaped to $c_t \times \frac{W}{32} \times \frac{H}{32}$. Finally, we connect the ASPP~\cite{chen2017deeplab} layer to extract global contextual information efficiently by increasing the receptive field of the model. We then feed the output of the ASPP layer $X _ { out }$ to the decoder.

\begin{table*}[t!]
	\centering
	\resizebox{2\columnwidth}{!}{%
		\begin{tabular}{r|c|ccc|cccc}
			\hline
			\multicolumn{1}{c|}{Method}  & cap & $\delta<1.25 \uparrow$ & $\delta<1.25^2 \uparrow$ & $\delta<1.25^3 \uparrow$ & AbsRel $\downarrow$ & SqRel $\downarrow$ & RMSE $\downarrow$ & RMSE \textit{log} $\downarrow$ \\ \hline \hline
			
			Eigen et al. \cite{eigen2014depth}                 & 0-80m & 0.702 & 0.898 & 0.967 & 0.203 & 1.548 & 6.307 & 0.282 \\
			Godard et al. \cite{godard2017unsupervised} & 0-80m & 0.861 & 0.949 & 0.976 & 0.114 & 0.898 & 4.935 & 0.206 \\
			Kuznietsov et al. \cite{kuznietsov2017semi}        & 0-80m & 0.862 & 0.960 & 0.986 & 0.113 & 0.741 & 4.621 & 0.189 \\
			Gan et al. \cite{gan2018monocular}                 & 0-80m & 0.890 & 0.964 & 0.985 & 0.098 & 0.666 & 3.933 & 0.173 \\
			Fu et al. \cite{fu2018deep}                        & 0-80m & 0.932 & 0.984 & 0.994 & 0.072 & 0.307 & 2.727 & 0.120 \\ 
			Yin et al. \cite{yin2019enforcing}                 & 0-80m & 0.938 & 0.990 & \underline{0.998} & 0.072 &   -   & 3.258 & 0.117 \\ 
			Lee et al. \cite{lee2019big}                       & 0-80m & \underline{0.956} & \underline{0.993} & \underline{0.998} & \bf{0.059} & 0.245 & 2.756 & \underline{0.096} \\ \hline
			
			$\star$ MobileNetV2~\cite{sandler2018mobilenetv2}                                        & 0-80m & 0.935 & 0.991 & \underline{0.998} & 0.071 & 0.291 & 2.970 & 0.110 \\
			$\star$ Densenet161~\cite{huang2017densely}                                      & 0-80m & 0.944 & 0.992 & \underline{0.998} & 0.069 & 0.272 & 2.829 &  0.099 \\
			$\star$ EfficientNet-B6~\cite{tan2019efficientnet}                                       & 0-80m & 0.950 & 0.992 & \underline{0.998} & 0.067 & 0.254 & 2.785 & 0.098 \\ \hline
			
			Ours-MobileNetV2                                        & 0-80m & 0.947 & 0.992 & \underline{0.998} & 0.065 & 0.261 & 2.925 & 0.107 \\
			Ours-Densenet161                                      & 0-80m & \underline{0.956} & \bf{0.994} & \bf{0.999} & 0.063 & \underline{0.243} & \underline{2.711} & \underline{0.096} \\
			Ours-EfficientNet-B6                                       & 0-80m & \bf{0.958} & \bf{0.994} & \bf{0.999} & \underline{0.060} & \bf{0.231} & \bf{2.642} & \bf{0.094} \\ \hline \hline
			
			Garg et al. \cite{garg2016unsupervised}            & 0-50m & 0.740 & 0.904 & 0.962 & 0.169 & 1.080 & 5.104 & 0.273 \\
			Godard et al. \cite{godard2017unsupervised} & 0-50m & 0.873 & 0.954 & 0.979 & 0.108 & 0.657 & 3.729 & 0.194 \\
			Kuznietsov et al. \cite{kuznietsov2017semi}        & 0-50m & 0.875 & 0.964 & 0.988 & 0.108 & 0.595 & 3.518 & 0.179 \\
			Gan et al. \cite{gan2018monocular}                 & 0-50m & 0.898 & 0.967 & 0.986 & 0.094 & 0.552 & 3.133 & 0.165 \\
			Fu et al. \cite{fu2018deep}                        & 0-50m & 0.936 & 0.985 & 0.995 & 0.071 & 0.268 & 2.271 & 0.116 \\
			Lee et al. \cite{lee2019big}                       & 0-50m & \bf{0.964} & \underline{0.994} & \bf{0.999} & \bf{0.056} & \underline{0.169} & \underline{1.925} & \bf{0.087} \\ \hline
			
			$\star$ MobileNetV2~\cite{sandler2018mobilenetv2}                                        & 0-50m & 0.943 & 0.992 & \underline{0.998} & 0.070 & 0.253 & 2.539 & 0.101 \\
			$\star$ Densenet161~\cite{huang2017densely}                                      & 0-50m & 0.950 & 0.993 & \underline{0.998} & 0.067 & 0.202 & 2.014 &  0.091 \\
			$\star$ EfficientNet-B6~\cite{tan2019efficientnet}                                       & 0-50m & 0.958 & 0.993 & \underline{0.998} & 0.061 & 0.182 & 1.993 & 0.089 \\ \hline
			
			Ours-MobileNetV2                                        
			& 0-50m & 0.958 & 0.993 & \underline{0.998} & 0.068 & 0.221 & 2.127 & 0.098 \\			
			Ours-Densenet161                                      
			& 0-50m & \underline{0.963} & \underline{0.994} & \bf{0.999} & 0.060 & 0.177 & 1.960 & 0.088 \\
			Ours-EfficientNet-B6                                         
			& 0-50m & \bf{0.964} & \bf{0.995} & \bf{0.999} & \underline{0.057} & \bf{0.168} & \bf{1.897} & \bf{0.087} \\ \hline
		\end{tabular}
	}
	\vspace{0.5cm}
	\caption{Performance comparison with other state-of-the-art methods on the KITTI Eigen split. $\uparrow$ indicates that higher is better and $\downarrow$ indicates that lower is better. Baseline models are marked with $\star$ symbol. The top two results are marked with \textbf{bold} and \underline{underline}. The same notation is used in the following sections. All methods were evaluated on the split given by Eigen \emph{et~al}.~\cite{eigen2014depth}. Our approach achieved state-of-the-art results.}
	\label{tb:1}	
\end{table*}

\begin{table}[t]
	\centering
	\resizebox{1\columnwidth}{!}{%
		\begin{tabular}{r|ccc|ccc}
			\hline
			\multicolumn{1}{c|}{Method}       & $\delta<1.25 \uparrow$ & $\delta<1.25^2 \uparrow$ & $\delta<1.25^3 \uparrow$ & AbsRel $\downarrow$ & RMSE $\downarrow$ & log10 $\downarrow$\\ \hline \hline
			Eigen et al. \cite{eigen2015predicting} & 0.769 & 0.950 & 0.988 & 0.158 & 0.641 & \multicolumn{1}{c}{-} \\
			Chakrabarti et al. \cite{chakrabarti2016depth} & 0.806 & 0.958 & 0.987 & 0.149 & 0.620 & \multicolumn{1}{c}{-} \\
			Li et al. \cite{li2017two}   & 0.789 & 0.955 & 0.988 & 0.152 & 0.611 & 0.064                  \\
			Xu et al. \cite{xu2017multi}   & 0.811 & 0.954 & 0.987 & 0.121 & 0.586 & 0.052                  \\
			Lee at al. \cite{lee2018single}   & 0.815 & 0.963 & 0.991 & 0.139 & 0.572 & \multicolumn{1}{c}{-} \\
			Qi et al. \cite{qi2018geonet}    & 0.834 & 0.960 & 0.990 & 0.128 & 0.569 & 0.057 \\ 
			Hu et al. \cite{hu2019revisiting} & 0.866 & 0.975 & 0.993 & 0.115 & 0.530 & 0.050 \\ 
			Chen et al. \cite{chen2019structure} & 0.878 & 0.977 & 0.994 & 0.111 & 0.514 & 0.048 \\
			Yin et al. \cite{yin2019enforcing} & 0.875 & 0.976 & 0.994 & 0.108 & 0.416 & 0.048 \\
			Lee et al. \cite{lee2019big} & 0.885 & 0.978 & 0.994 & 0.110 & 0.392 & 0.047 \\
			Chen et al. \cite{chen2020improving} & \bf{0.899} & \underline{0.983} & \underline{0.996} & \bf{0.098} & \underline{0.376} & \bf{0.042} \\ \hline
			
			$\star$ MobileNetV2~\cite{sandler2018mobilenetv2} & 0.805 & 0.962 & 0.991 & 0.149 & 0.488 & 0.062 \\
			$\star$ Densenet161~\cite{huang2017densely} & 0.862 & 0.975 & 0.994 & 0.122 & 0.416 & 0.051 \\
			$\star$ EfficientNet-B6~\cite{tan2019efficientnet} & 0.871 & 0.981 & \underline{0.996} & 0.115 & 0.399 & 0.049 \\ \hline
			
			Ours-MobileNetV2   & 0.838 & 0.969 & 0.993 & 0.135 & 0.439 & 0.055 \\
			Ours-Densenet161  & 0.881 & 0.981 & \underline{0.996} & 0.113 & 0.387 & 0.047 \\
			Ours-EfficientNet-B6 & \underline{0.893} & \bf{0.985} & \bf{0.997} & \underline{0.107} & \bf{0.373} & \underline{0.046} \\ \hline
		\end{tabular}
	}
	\vspace{0.5cm}
	\caption{Performance comparison with other state-of-the-art methods on the NYU Depth V2 test set.}
	\label{tb:2}	
\end{table}

\subsection{Objective Functions}
We apply the scale-invariant error proposed by Eigen \emph{et~al}.~\cite{eigen2014depth} as the training loss. The objective function is defined as follows:

\begin{equation} 
	L= \sqrt { \frac{ 1 } { T } \sum _ { i } ^ { } g _ { i } ^ { 2 } - \frac{ \lambda } { T ^ { 2 } } \left ( \sum _ { i } ^ { } g _ { i } \right ) ^ { 2 } } ,
\end{equation}

\noindent
where $g _ { i } =\log d _ { i } -\log d _ { i } ^ { \ast }$ with the ground truth depth $d_{i}$ and the predicted depth $d _ { i } ^ { \ast }$. $T$ is the number of valid pixels of the ground truth. Specifically, the loss is a mixture of elementwise $l_2$ and the scale-invariant error. Because it is the sum of the variance and the weighted squared mean of the error in log space, setting a high $\lambda$ enforces a greater focus on minimizing the variance of the error. Therefore, we set $\lambda$ to 0.85, the same as in~\cite{lee2019big}.


\section{Experiments}
\subsection{Datasets and Evaluation Metrics}
\noindent
\textbf{NYU Depth V2 Dataset.} The NYU Depth V2 dataset~\cite{silberman2012indoor} contains 464 indoor scenes, consisting of 120K images and paired depth maps with a resolution of $640 \times 480$ pixels. We followed the official train/test split from previous studies, using 249 scenes for training and 215 scenes (654 images) for testing. Due to the asynchronous capture rates between the original RGB images and the depth maps, all the images and depth maps used for the experiments on NYU Depth V2 were collected from Lee \emph{et~al}.~\cite{lee2019big}. In the evaluation, we applied a predefined center cropping method by Eigen \emph{et~al}.~\cite{eigen2014depth}.

\begin{figure*}[t]
	\setlength{\belowcaptionskip}{-10pt}
	\begin{center}
		\resizebox{1.8\columnwidth}{!}{%
			\includegraphics[width=1\linewidth]{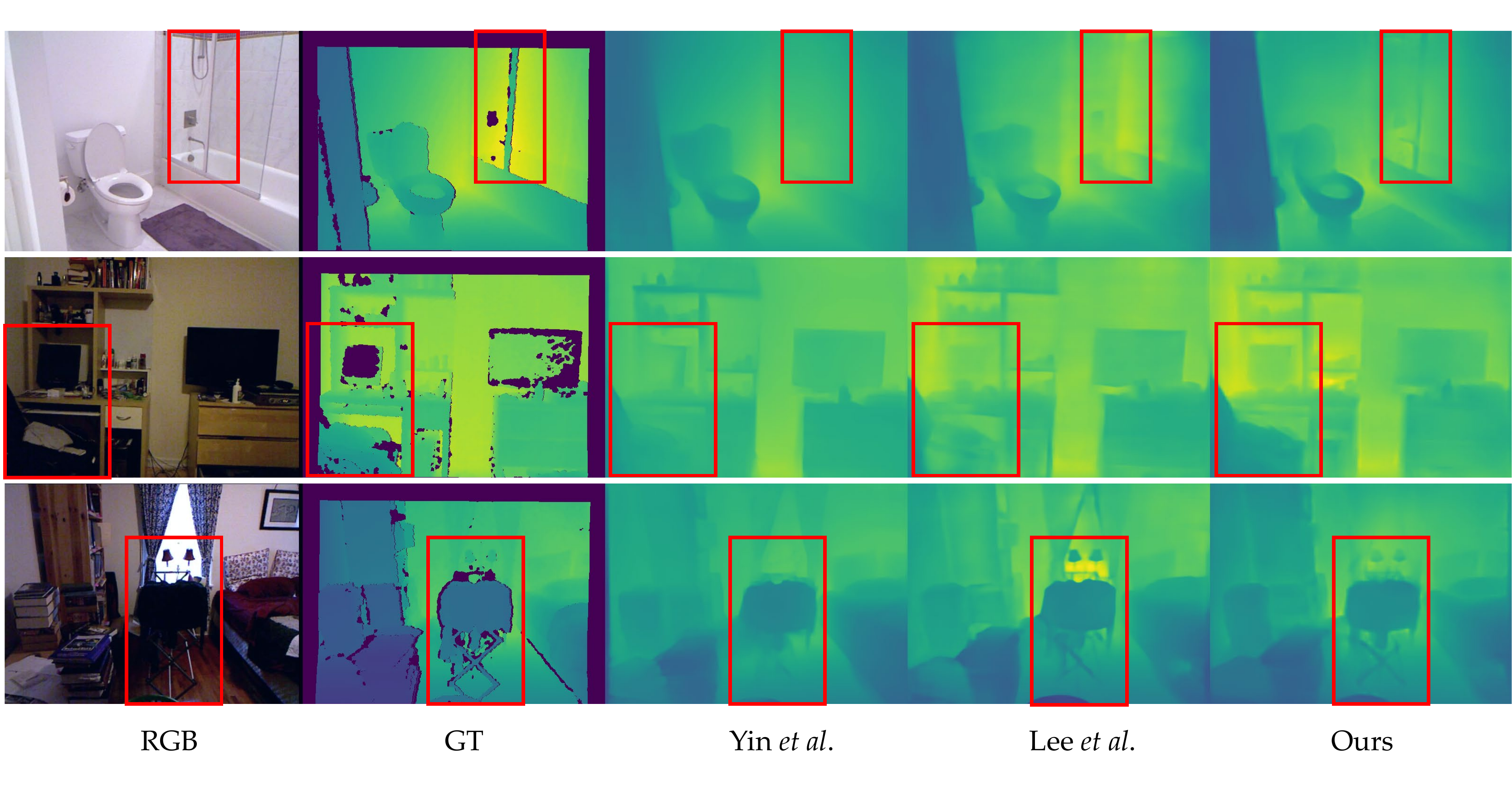}
		}
	\end{center}
	\vspace{-0.5cm}
	\caption{Comparison of qualitative results for the NYU Depth V2 test set with other state-of-the-art methods by Yin \emph{et~al}.~\cite{yin2019enforcing} and Lee \emph{et~al}.~\cite{lee2019big}. Our method is based on the EfficientNet-B6 encoder.}
	\label{fig:4}
\end{figure*}

\noindent
\textbf{KITTI Dataset.} The KITTI dataset~\cite{geiger2013vision} contains scenes from 61 outdoor driving scenarios captured by multiple sensors. The dataset contains RGB images with a resolution of $1241 \times 375$ pixels and the corresponding LiDAR point clouds. We used the official KITTI dataset containing RGB images and post-processed depth maps of projected LiDAR point clouds as the ground truth. The ground truth is produced by combining LiDAR scans. In the experiments, we followed the common data split proposed by Eigen \emph{et~al}.~\cite{eigen2014depth} for comparison with previous studies. Therefore, 697 images covering a total of 29 scenes were used for evaluation, and the remaining 23,488 images covering 32 scenes were used for training. For the KITTI dataset, we trained our networks on $1216 \times 352$ pixels using a bottom-center crop.

\noindent
\textbf{Evaluation Metrics.} Following the previous work~\cite{eigen2014depth}, we evaluate our methods using the following quantitative metrics:

\begin{itemize}
	\item Accuracy with threshold: percentage (\%) of $d_{i}$ s.t. $max\left(\dfrac{d_{i}}{d_{i}^{*}},\dfrac{d_{i}^{*}}{d_{i}}\right)=\delta<thr,thr=1.25,1.25^{2},1.25^{3}$
	
	\item RMSE : $\sqrt{\dfrac{1}{N}\sum_{i}^N \left ( d_{i}-d_{i}^{*} \right )^{2}}$
	
	\item RMSE \textit{log} : $\sqrt{\dfrac{1}{N}\sum_{i}^N||\log_{10} d_{i}-\log_{10} d_{i}^{*}||^{2}}$
	
	\item AbsRel: $\dfrac{1}{N}\sum_{i}^N \dfrac{|d_{i}-d_{i}^{*}|}{d_{i}^{*}}$
	
	\item SqRel: $\dfrac{1}{N}\sum_{i}^N \dfrac{||d_{i}-d_{i}^{*}||^{2}}{d_{i}^{*}}$
\end{itemize}

\noindent
, where $d_{i}$ and $d_{i}^{*}$ is the ground-truth depth and the estimated depth at pixel $i$, respectively; $N$ denotes a collection of pixels that the ground truth values are available. In addition, we used mean log10 error (log10) for comparison with~\cite{lee2019big}.

\subsection{Implementation Details}
\label{sect:4.2}
We implemented the proposed method using the open deep learning framework PyTorch. For network training, we used the Adam optimizer~\cite{kingma2014adam} with $\beta_1=0.9$, $\beta_2=0.999$, and $\epsilon =10 ^ { -6 }$. The learning rate decayed from $10^{-4}$ to $10^{-5}$ with the cosine annealing scheduler~\cite{loshchilov2016sgdr}. The total number of epochs was set to 50 with batch size 4 with a single NVIDIA Quadro RTX 6000 GPU for all experiments in this study.

For the NYU Depth V2 dataset, we set the patch sizes of the patch embedding functions $f_e$ in the two PEMs to $4 \times 4$ and $2 \times 2$. Because each input feature map resolution was $80 \times 60$ and $40 \times 30$, the number of patches $N_p$ generated by each patch embedding function was 300. For the KITTI dataset, the size of each PEM's input feature map was $152 \times 44$ and $76 \times 22$. Because we applied the same patch size to the KITTI dataset, the number of patches was 418.

We used EfficientNet-B6~\cite{tan2019efficientnet}, DenseNet161~\cite{huang2017densely}, and MobileNetV2~\cite{sandler2018mobilenetv2}, which were pretrained on image classification using the ImageNet-1K dataset~\cite{russakovsky2015imagenet}, as the backbone networks. To avoid overfitting, we applied data augmentation to all the experiments. Random horizontal flipping was applied in all cases. We also randomly rotated the input images in ranges of $\left[ -1,1 \right]$ and $\left[ -2.5,2.5 \right]$ degrees for the KITTI and NYU Depth V2 datasets, respectively. Thereafter, contrast, brightness, and color adjustment were applied to the inputs for both datasets.

\begin{figure*}[t]
	\setlength{\belowcaptionskip}{-10pt}
	\begin{center}
		\includegraphics[width=1\linewidth]{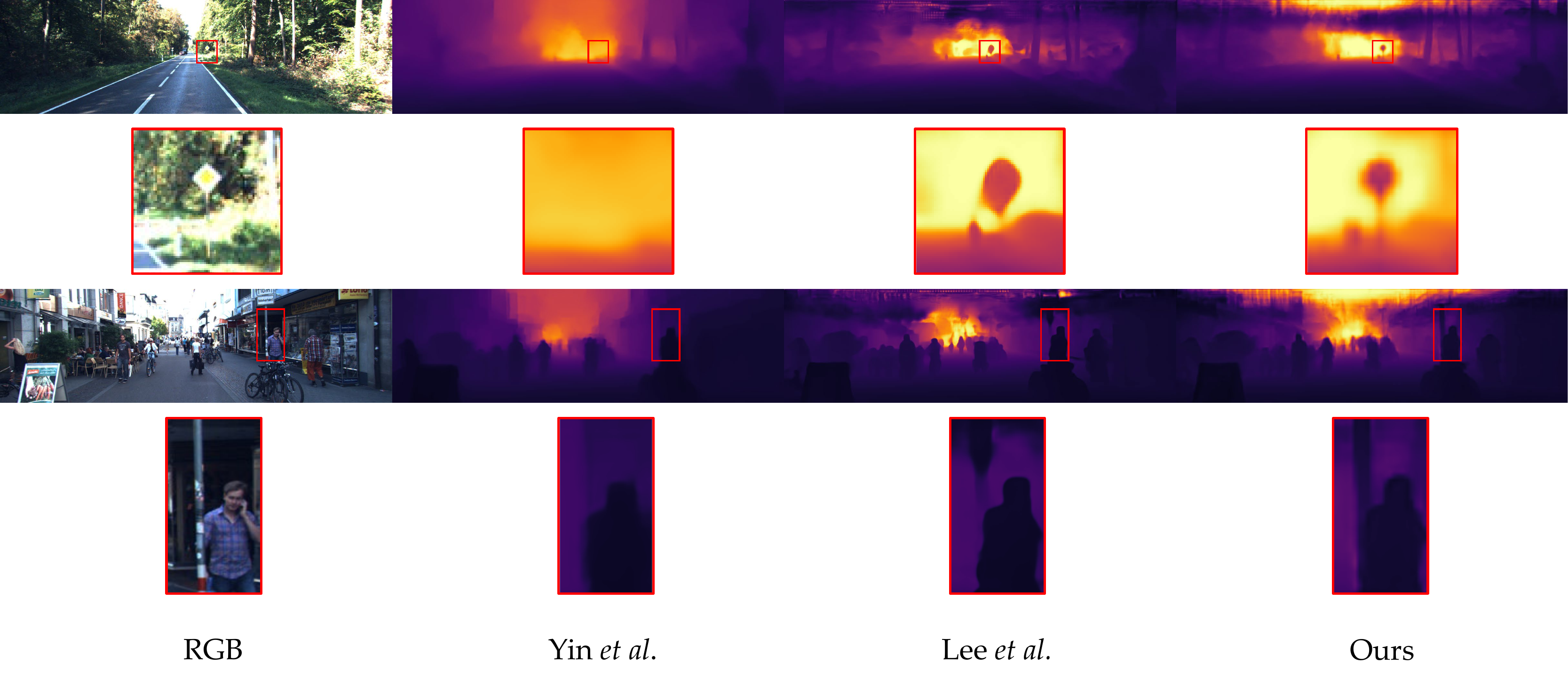}
	\end{center}
	\vspace{-0.5cm}
	\caption{Comparison of qualitative results for the  KITTI Eigen split with other state-of-the-art methods by Yin \emph{et~al}.~\cite{yin2019enforcing} and Lee \emph{et~al}.~\cite{lee2019big}. Our method is based on the EfficientNet-B6 encoder.}
	\vspace{-0.5cm}
	\label{fig:5}
\end{figure*}

\subsection{Results}
Tables~\ref{tb:1} and \ref{tb:2} compare the performance results of the proposed method with previously presented state-of-the-art algorithms for the NYU Depth V2 and KITTI datasets. The proposed method was evaluated with the backbone encoders MobileNetV2~\cite{sandler2018mobilenetv2}, DenseNet161~\cite{huang2017densely}, and EfficientNet-B6~\cite{tan2019efficientnet}. All these models used PEM and EAM, as shown in Figure~\ref{fig:2}, and each performance effect is described in detail in the ablation study section. 

\noindent
\textbf{NYU Depth V2 Dataset.} As shown in Table~\ref{tb:2}, our EfficientNet-B6 based model outperforms the previous methods~\cite{eigen2015predicting, chakrabarti2016depth, li2017two, xu2017multi, lee2018single, qi2018geonet, hu2019revisiting, chen2019structure, yin2019enforcing, lee2019big, chen2020improving} in the NYU Depth V2 dataset. In particular, the RMSE was the lowest, at 0.373. RMSE measures depth errors directly and applies a heavier penalty for undesirable larger errors. Therefore, RMSE is known as the main evaluation metric of depth estimation.

\begin{figure}[t]
	\setlength{\belowcaptionskip}{-10pt}
	\begin{center}
		\includegraphics[width=1\linewidth]{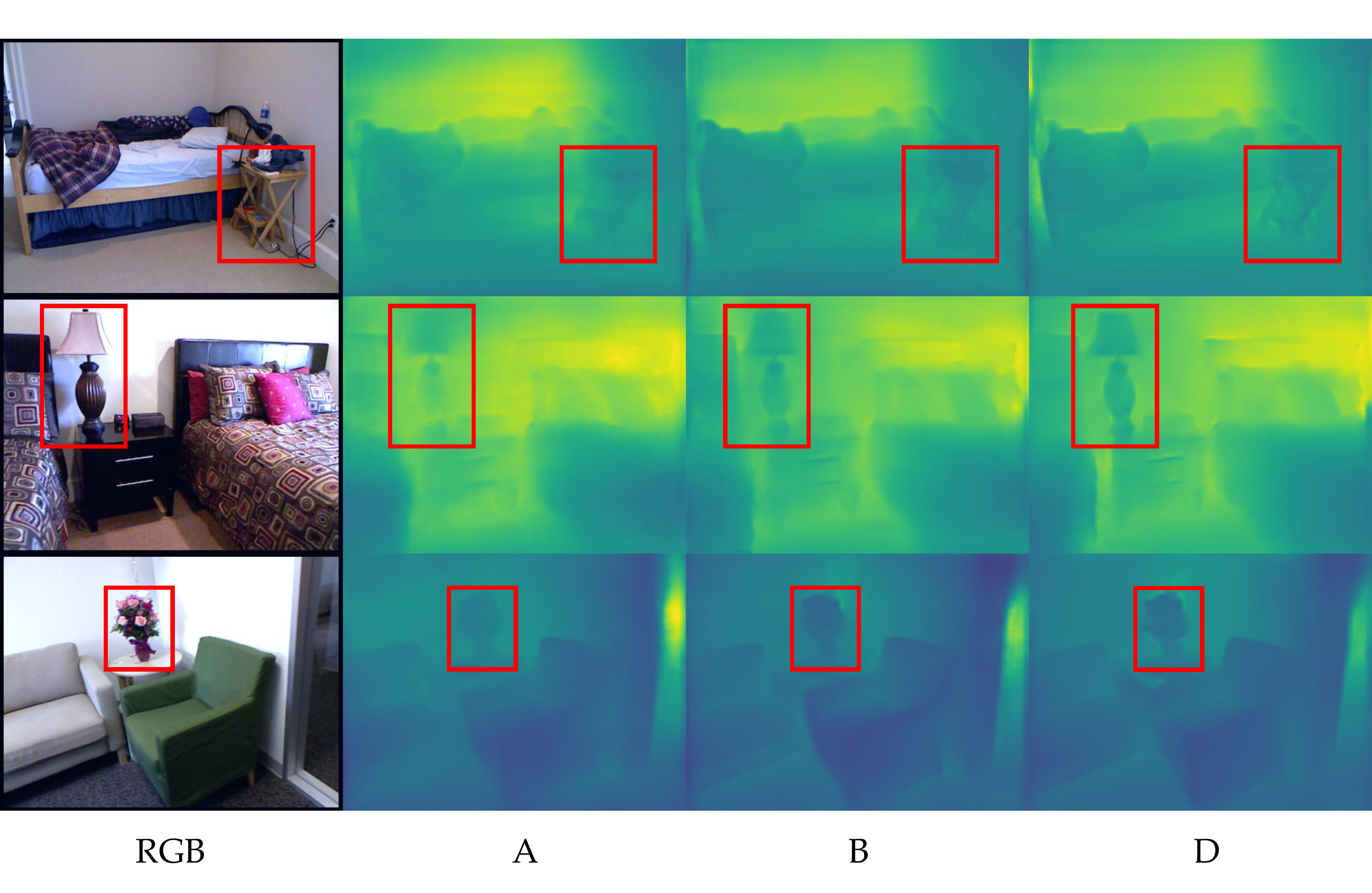}
	\end{center}
	\vspace{-0.5cm}
	\caption{Comparison of qualitative results for the NYU Depth V2 test set using the baseline model (A), baseline with PEM (B), and the proposed PEM and EAM combination (D). A, B, and D are the model names shown in Table~\ref{tb:3}. The baseline model is MobileNetV2.}
	\label{fig:6}
\end{figure}

The NYU Depth V2 dataset is an indoor dataset containing a large amount of furniture with complex surface structures. Moreover, light reflections and shadows caused by windows and glass make it difficult to predict the depth map. Therefore, to predict the depth of areas lacking visual cues, it is important for the model to extract structural information from the relationships between adjacent pixels. Figure~\ref{fig:4} shows the qualitative results of our method for various challenging indoor scenes. The results in Figure~\ref{fig:4} show that our method robustly predicts depth in extreme lighting conditions. In particular, our model generates accurate depth maps in low-light environments such as the second row and intense lighting conditions such as the third row of Figure~\ref{fig:4}. Furthermore, the first row of Figure~\ref{fig:4} shows the results of our method in an optical illusion situation caused by an obstacle on the glass. However, the proposed PEM and EAM help the model reinforce the learning of relationships between neighboring patches to make accurate depth prediction possible.


\begin{table*}[t]
	\centering
	\resizebox{1.8\columnwidth}{!}{%
		\begin{tabular}{c|ccc|ccc|ccc}
			\hline
			name & Baseline & EAM & PEM & $\delta<1.25 \uparrow$ & $\delta<1.25^2 \uparrow$ & $\delta<1.25^3 \uparrow$ & AbsRel $\downarrow$ & RMSE $\downarrow$ & log10 $\downarrow$\\ \hline \hline
			A & \ding{51} & & & 0.805 & 0.962 & 0.991 & 0.149 & 0.488 & 0.062\\
			B & \ding{51} & & \ding{51} & 0.823 & 0.964 & 0.992 & 0.140 & 0.461 & 0.060 \\
			C & \ding{51} & \ding{51} & & 0.835 & 0.968 & 0.992 & 0.136 & 0.443 & 0.056 \\
			D & \ding{51} & \ding{51} & \ding{51} & \textbf{0.838} & \textbf{0.969} & \textbf{0.993} & \textbf{0.135} & \textbf{0.439} & \textbf{0.055} \\
			\hline
		\end{tabular}
	}
	\vspace{0.5cm}
	\caption{Evaluation results of the baseline model (MobileNetV2) and the proposed model with the PEM and EAM combination for the NYU test set.}
	\label{tb:3}	
\end{table*}

\begin{table*}[t]
	\centering
	\resizebox{2\columnwidth}{!}{%
		\begin{tabular}{c|ccc|c|ccc|cccc}
			\hline
			name & Baseline & EAM & PEM & cap & $\delta<1.25 \uparrow$ & $\delta<1.25^2 \uparrow$ & $\delta<1.25^3 \uparrow$ & AbsRel $\downarrow$ & SqRel $\downarrow$ & RMSE $\downarrow$ & RMSE \textit{log} $\downarrow$ \\ \hline \hline
			A & \ding{51} & & & 0-80m & 0.935 & 0.991 & \bf{0.998} & 0.071 & 0.291 & 2.970 & 0.110 \\
			B & \ding{51} & & \ding{51} & 0-80m & 0.939 & 0.991 & \textbf{0.998} & 0.068 & 0.273 & 2.956 & 0.109 \\
			C & \ding{51} & \ding{51} & & 0-80m & 0.946 & \textbf{0.992} & \bf{0.998} & \textbf{0.065} & 0.269 & 2.931 & \textbf{0.107} \\
			D & \ding{51} & \ding{51} & \ding{51} & 0-80m & \textbf{0.947} & \textbf{0.992} & \bf{0.998} & \textbf{0.065} & \textbf{0.261} & \textbf{2.925} & \textbf{0.107} \\
			\hline
		\end{tabular}
	}
	\vspace{0.5cm}
	\caption{Evaluation results of the baseline model (MobileNetV2) and the proposed model with the PEM and EAM combination for the KITTI Eigen split.}
	\label{tb:4}	
\end{table*}

\begin{figure*}[!t]
	\setlength{\belowcaptionskip}{-10pt}
	\begin{center}
		\includegraphics[width=0.95\linewidth]{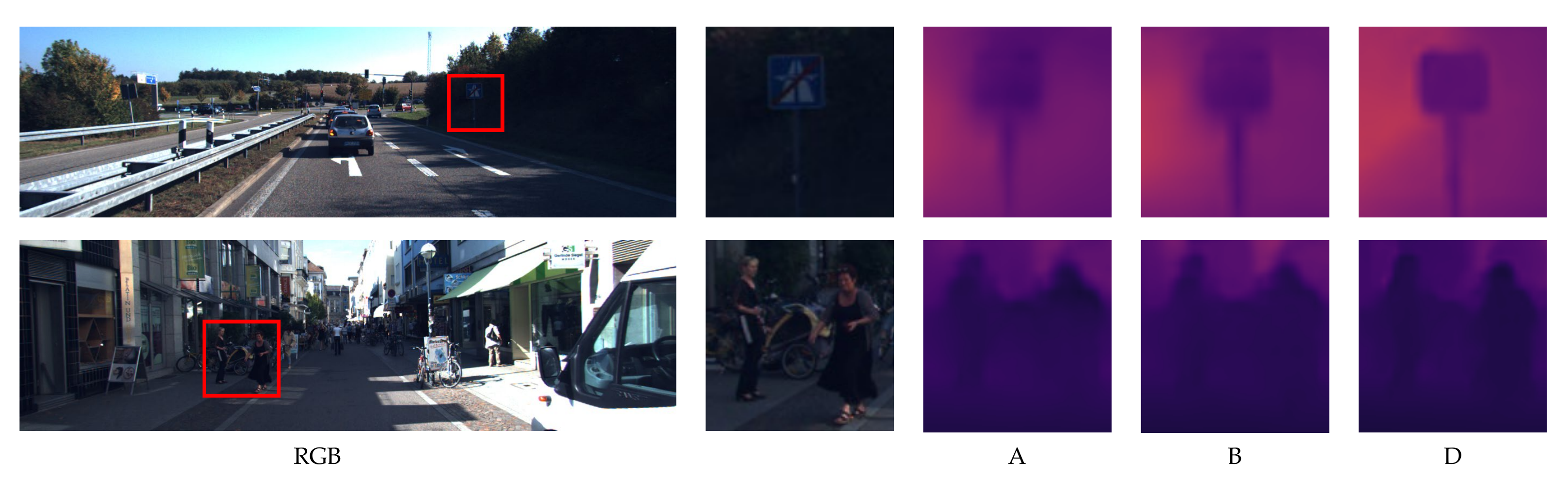}
	\end{center}
	\vspace{-0.5cm}
	\caption{Comparison of qualitative results of the baseline model (MobileNetV2), baseline with PEM, and the proposed model with the PEM and EAM combination for the KITTI Eigen split. A, B, and D are the model names shown in Table~\ref{tb:4}.}
	\label{fig:7}
\end{figure*}

\noindent
\textbf{KITTI dataset.} In general, outdoor scenes are more challenging because the lighting changes are more severe than in indoor scenes. The KITTI dataset contains images of outdoor driving scenarios for autonomous vehicles. These images include many trees and buildings with complex structures, along with traffic signs with thin columns, making it difficult to predict the depth. Table~\ref{tb:1} shows the performance of the proposed model for the KITTI dataset compared to the previous methods~\cite{eigen2014depth, godard2017unsupervised, kuznietsov2017semi, gan2018monocular, fu2018deep, yin2019enforcing, lee2019big, garg2016unsupervised}. As shown in Table~\ref{tb:1}, our EfficientNet-B6-based model achieved state-of-the-art performance in most evaluation metrics. The qualitative results in Figure~\ref{fig:5} support the standpoint that our model also robustly estimated depth for outdoor scenes. In particular, the proposed method accurately predicted depth information for long and thin objects such as traffic signs and traffic cones. Furthermore, our model produced results that were more robust to changes in brightness. This is because the PEM and EAM extract structural information from the relationships between patches, and thus they supplement the lack of visual cues with related adjacent patches. We detail the effects of the proposed PEM and EAM in the next section.

\noindent
\textbf{Impact of the PEM and EAM.}
We performed an ablation study on the NYU Depth V2 and KITTI datasets to further demonstrate the impact of the proposed PEM and EAM. The baseline model encoder for the ablation study was MobileNetV2, and the implementation details of all the experiments were the same as in Section~\ref{sect:4.2}. Tables~\ref{tb:3} and~\ref{tb:4} show our experimental results for the NYU Depth V2 and KITTI datasets, respectively. As shown in Tables~\ref{tb:3} and~\ref{tb:4}, the model has the highest depth estimation performance when the PEM and the EAM were all used. The EAM, including EdgeConv, learns structure information by extracting edge features between patches, like the PEM. However, using the PEM, the model can learn more diverse inter-patch relationships from multi-scale feature maps. Therefore, the performance was further improved with the PEM. Figures~\ref{fig:6} and~\ref{fig:7} show the qualitative results of the ablation studies. Our model more accurately estimated the depth map and generated sharp object contours than the baseline model. This is because both global contextual information and structural information are important in monocular depth estimation. The experimental results support that the proposed model learned structural information by learning the relationships between nearby pixels.

\section{Conclusion}
In this paper, we propose the novel PEM and EAM for monocular depth prediction. We apply EdgeConv for depth prediction to extract edge features between patches close to each other in the feature space. The proposed method makes robust depth prediction by learning structural information through edge features. Our model achieves state-of-the-art performance on NYU Depth V2 and KITTI datasets, especially with the smallest RMSE. The ablation study proves the effectiveness of the proposed modules and shows that the depth prediction performance is greatly improved in challenging scenes.

\noindent
\\
\footnotesize\textbf{Acknowledgement.} This work was supported by the Institute of Information \& communications Technology Planning \& Evaluation(IITP) grant funded by the Korea government(MSIT) (No. 2021-0-00172, The development of human Re-identification and masked face recognition based on CCTV camera)

\newpage

{\small
\bibliographystyle{ieee_fullname}
\bibliography{egbib}
}

\end{document}